\documentclass[pdflatex,sn-mathphys]{sn-jnl}% Math and Physical Sciences Reference Style
\usepackage{lipsum}
\usepackage{lscape}
\jyear{2021}%

\theoremstyle{thmstyleone}%
\theoremstyle{thmstyletwo}%

\theoremstyle{thmstylethree}%

\begin{document}
\vspace{-10cm}
\title[Non-invasive hemodynamic analysis for aortic regurgitation]{\textbf{Non-invasive hemodynamic analysis for aortic regurgitation using computational fluid dynamics and deep learning}}


\author*[1]{\fnm{Derek} \sur{Long}}\email{dlon450@aucklanduni.ac.nz}

\author[1]{\fnm{Cameron} \sur{McMurdo}}\email{cmcm658@aucklanduni.ac.nz}

\author[2]{\fnm{Edward} \sur{Ferdian}}\email{e.ferdian@aucklanduni.ac.nz}

\author[3]{\fnm{Charl\`ene A.} \sur{Mauger}}\email{c.mauger@auckland.ac.nz}

\author[4,5]{\fnm{David} \sur{Marlevi}}\email{marlevi@mit.edu}

\author[2,6]{\fnm{Alistair A.} \sur{Young}}\email{a.young@auckland.ac.nz}

\author[1,3]{\fnm{Martyn P.} \sur{Nash}}\email{martyn.nash@auckland.ac.nz}

\affil[1]{\orgdiv{Department of Engineering Science}, \orgname{University of Auckland}, \orgaddress{\city{Auckland}, \country{New Zealand}}}

\affil[2]{\orgdiv{Department of Anatomy and Medical Imaging}, \orgname{University of Auckland}, \orgaddress{\city{Auckland}, \country{New Zealand}}}

\affil[3]{\orgdiv{Auckland Bioengineering Institute}, \orgname{University of Auckland}, \orgaddress{\city{Auckland}, \country{New Zealand}}}

\affil[4]{\orgdiv{Institute for Medical Engineering and Science}, \orgname{Massachusetts Institute of Technology}, \orgaddress{\city{Cambridge}, \state{MA}, \country{USA}}}

\affil[5]{\orgdiv{Department of Molecular Medicine and Surgery}, \orgname{Karolinska Institutet}, \orgaddress{\city{Solna}, \country{Sweden}}}

\affil[6]{\orgdiv{Department of Biomedical Engineering}, \orgname{King's College London}, \orgaddress{\city{London}, \country{United Kingdom}}}


\abstract{Changes in cardiovascular hemodynamics are closely related to the development of aortic regurgitation, a type of valvular heart disease. Metrics derived from blood flows are used to indicate aortic regurgitation onset and evaluate its severity. These metrics can be non-invasively obtained using four-dimensional (4D) flow magnetic resonance imaging (MRI), where accuracy is primarily dependent on spatial resolution. However, insufficient resolution often results from limitations in 4D flow MRI and complex aortic regurgitation hemodynamics. To address this, computational fluid dynamics simulations were transformed into synthetic 4D flow MRI data and used to train a variety of neural networks. These networks generated super resolution, full-field phase images with an upsample factor of 4. Results showed decreased velocity error, high structural similarity scores, and improved learning capabilities from previous work. Further validation was performed on two sets of in-vivo 4D flow MRI data and demonstrated success in de-noising flow images. This approach presents an opportunity to comprehensively analyse aortic regurgitation hemodynamics in a non-invasive manner.}

\clearpage\maketitle
\thispagestyle{empty}
% \begin{center}
%     \date{\today}
% \end{center}
% \newpage

\pagestyle{plain}
\setcounter{page}{1}

\section{Introduction}\label{sec1}

Aortic valve regurgitation, or aortic regurgitation (AR), is a common type of valvular heart disease where the aortic valve does not close properly, causing reflux of blood from the aorta into the left ventricle \cite{Valvular-Heart-Disease}. This reflux of blood is known as the regurgitant jet. Diagnosis and severity of AR is determined by evaluation of flow metrics, for example peak velocity, pressure drop, and regurgitant volume.

Cardiovascular four-dimensional (4D) flow magnetic resonance imaging (MRI) is a novel imaging technique to quantify full-field blood flow velocities, providing a three-dimensional (3D) velocity field across a region of interest throughout the cardiac cycle. Currently, due to the small width of the regurgitant jet and the limited spatiotemporal resolution of 4D flow MRI, it fails to accurately capture the complex hemodynamics of AR. The combination of computational fluid dynamics (CFD) and deep learning with 4D flow MRI will help to generate higher resolution images and recover hemodynamic parameters lost in current MRI images, with a target upsample factor of 4. This work will extend what has already been completed in 4DFlowNet \cite{4DFlowNet} by using a wider range of flow characteristics to mimic AR, improving the data augmentation steps, and enhancing the artificial neural network with newer architecture structures.

\section{Background}\label{sec2}

\subsection{Aortic Regurgitation}\label{clinical_back}
AR occurs when the aortic valve does not close properly, causing blood to flow back into the left ventricle from the aorta. This forces the heart to work harder and pump more blood to the aorta, which can cause further heart problems in the future. AR also has varying levels of intensity, from trace or mild through moderate to severe \cite{Valvular-Heart-Disease}. Acute AR is considered a medical emergency as it can cause severe pulmonary edema and hypotension, that is, excess fluid in the lungs and low blood pressure, respectively. Patients with AR are monitored yearly with echocardiography to decide whether replacement of the aortic valve is necessary \cite{Outcomes-after-aortic-valve}. Flow metrics such as peak velocity and pressure drop are typically calculated non-invasively using 2D velocities from 2D Doppler echocardiography \cite{Valvular-Heart-Disease}. Due to limited information available in 2D, this method is known to overestimate pressure drops \cite{marlevi2019estimation}.

\subsection{4D Flow MRI}\label{4dflowmri_lit}
4D flow MRI is an established imaging technique that captures the temporal changes of 3D blood flow patterns within individual vascular structures \cite{4DFLOWMRI, Itatani2017NewIT}. Velocities of blood particles are encoded in the phase of the MRI signal while the anatomy is visualised from the signal’s magnitude \cite{Jones98}. However, 4D flow has several limitations, such as low spatiotemporal resolution, long scan time, and low signal-to-noise (SNR) ratio \cite{JIANG2015185}, which makes its clinical application to AR difficult. With a spatial resolution between 1.0 and 3.5mm \cite{Itatani2017NewIT}, details on the narrowest part of the jet cannot be captured as its width is typically much smaller than 3mm \cite{Sallach2007} for mild AR. Therefore, spatial resolution is the biggest limitation in 4D flow MRI.

\subsection{Deep Learning in 4D Flow MRI}
Deep learning \cite{Deep-Learning} has had a significant impact in many scientific sectors, and is highly relevant in the field of medical imaging \cite{Medical-Image-DL}. Advances in super resolution image reconstruction \cite{SR-Image-Reconstruction} to obtain high-resolution (HR) images from low-resolution (LR) observations are increasingly being adopted for MRI with a deep learning-based approach \cite{MRI-SR-GAN-DenseNet, Brain-MRI-SR}. This approach is preferred as it not only has an advantage in spatial resolution quality over conventional super resolution techniques \cite{MRI-SR-CascadedDL}, but also successfully denoises flow images \cite{Cerebrovascular-4DFlowMRI}. 

The combination of 4D flow MRI with deep learning has been explored in multiple ways to increase resolution and provide more accurate estimates of physical quantities \cite{4DFlowNet,Physics-informed-4DFlowMRI,Ferdian-Cerebrov}. However, there are several limitations involved in this approach. Primarily, these have been related to insufficient data \cite{4DFlowNet,Ferdian-Cerebrov}, which is due to the requirement of paired LR and HR MR images. This can be difficult to obtain as HR MRI takes long scanning times and is subject to motion artifacts \cite{4DFlowNet,Ferdian-Cerebrov,Applications-DL-AA}. As an alternative, CFD models have been used to simulate 4D flow MRI as ground truth HR images, which are then downsampled to LR images \cite{4DFlowNet,Ferdian-Cerebrov}. Other limitations include unstable and non-robust network architectures \cite{Physics-informed-4DFlowMRI}, which describe the organisational structure of the network’s layers. The architectures plays a significant role in the performance of the DL algorithm, as well as ignoring phase/velocity aliasing error \cite{Cerebrovascular-4DFlowMRI, 4DFlowNet}. The aliasing error here refers to aliasing from having a velocity encoding (VENC) \cite{velocity-encoding} that is too low \cite{Aliasing-Encoding} rather than other types of MRI spatial aliasing which have been explored previously and reduced \cite{DL-AntiAliasing-MRI-SR, SMORE-anti-aliasing, Brain-MRI-SR-old}. Note that VENC is an MR parameter to adjust the maximum velocity corresponding to a 360$^{\circ}$ phase shift in the data.

\subsection{Network Architecture}
Recent development in object detection has proven significant in advancing ANN architecture \cite{Survey}, and appears to be widely used in many medical imaging applications \cite{Medical-Image-DL}. Examples include residual blocks \cite{ResNet}, dense blocks \cite{DenseNet}, and cross stage partial blocks \cite{CSPNet}, which show promise to increase network capacity mitigating degradation and memory utilisation issues.

\newpage

\section{Methods}\label{sec3}

\subsection{Data Generation}
Modelling of the aortic valve was done using Ansys 2021 R1, which has two main options for CFD simulations; CFX and Fluent. CFX was chosen as it is better suited for more simple, low Mach number flows, such as the flow through the aortic valve. Modelling was an iterative process to determine the best parameters and options that would increase efficiency and accuracy.

\subsubsection{Geometries}
\begin{figure}
    \centering
    \includegraphics[width=0.8\textwidth]{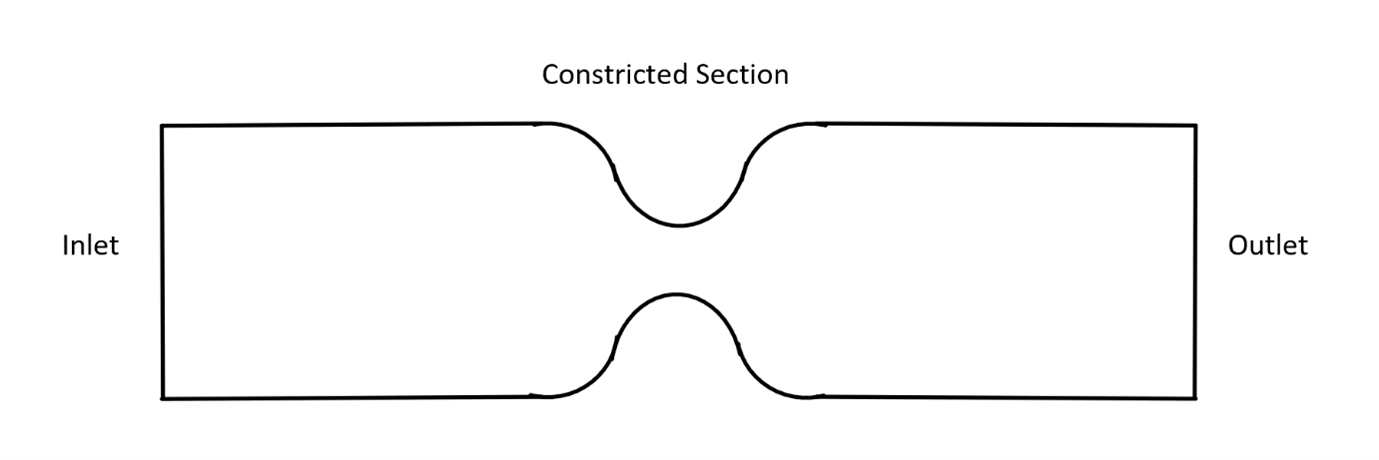}
    \caption{Basic sketch of the first 10 geometries.}
    \label{fig:basic-geom}
\end{figure}

The design of the problem geometry focused on simplicity over replicating a real-life aortic valve, allowing for geometries to be created easily. The basic geometry is shaped like a cylinder with a constricted section part way along the length, resembling a Venturi tube. The constricted section represents the gap in the aortiv valve present for aortic regurgitation to occur. Figure \ref{fig:basic-geom} shows a sketch of the basic geometry.

\begin{figure}
    \centering
    \includegraphics[width=0.8\textwidth]{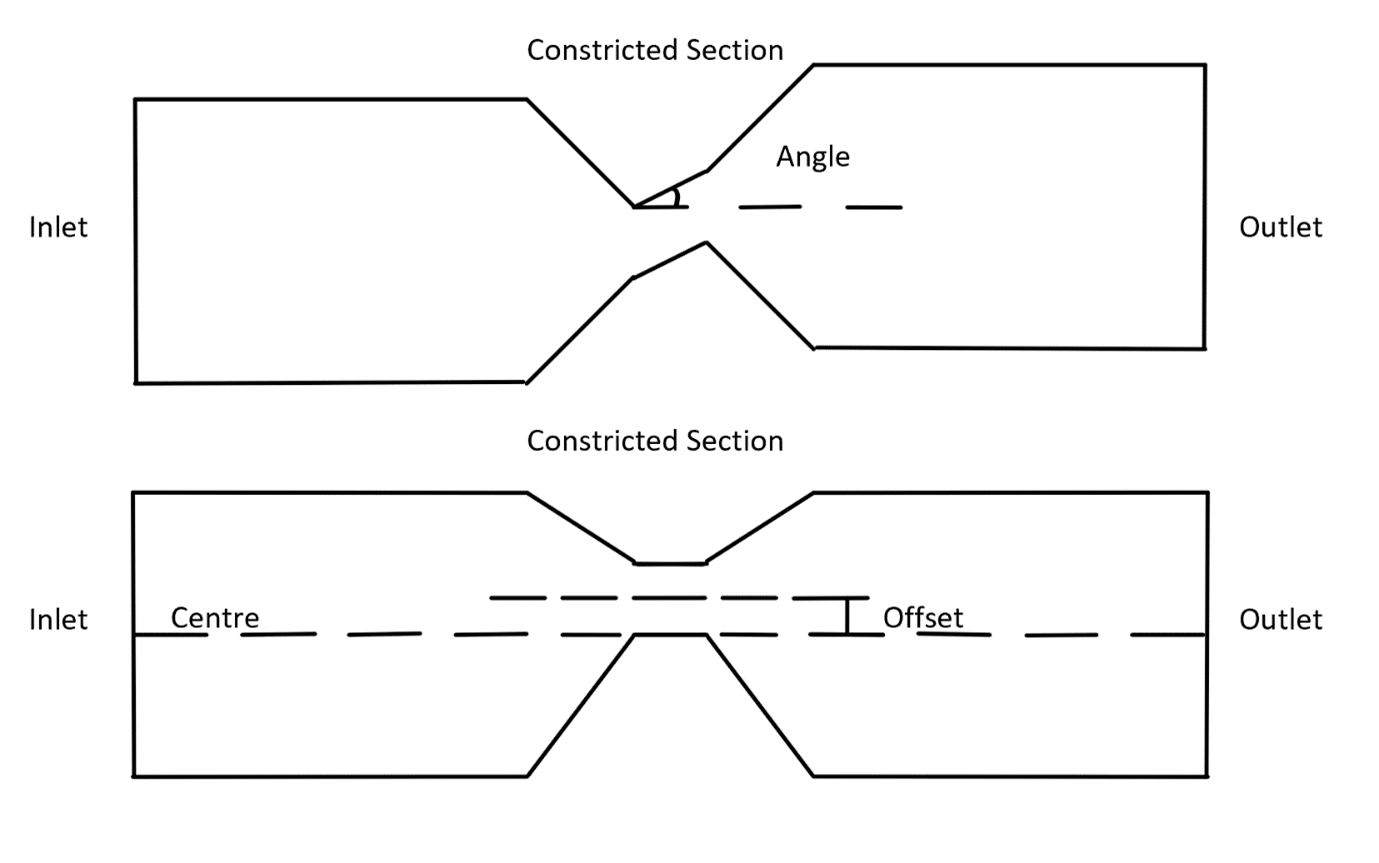}
    \caption{Sketches of the angled (top) and offset (bottom) constricted sections, respectively.}
    \label{fig:geom-1120}
\end{figure}

In total, 20 different geometries were generated. The first set of 10 geometries had variations in inlet velocity, inlet radius, and constricted section radius, while the other set of 10 geometries had different shapes. These geometries were designed to capture maximum jet velocities between 2.5 and 5.0 ms$^{-1}$. The second set of geometries were based on the third geometry, as this geometry was reasonably small with minimal computational time. This set had diagonal and off-centre constricted sections, as well combinations of both, to model eccentric jets. Figure \ref{fig:geom-1120} demonstrates these differences in shape. The parameters for all geometries are laid out in Table \ref{parameters}.

\begin{table}
\begin{center}
\begin{minipage}{\textwidth}
\caption{Input parameters for all geometries}\label{parameters}
\begin{tabular*}{\textwidth}{@{\extracolsep{\fill}}lrrrlrrr@{\extracolsep{\fill}}}
\toprule%%
\multicolumn{4}{@{}c@{}}{Basic Geometries\footnotemark[1]} & \multicolumn{4}{@{}c@{}}{Angled/Offset Geometries\footnotemark[2]} \\\cmidrule{1-4}\cmidrule{5-8}%
No. & $v_{max}$ & R$_I$ & R$_C$ & No. & $\theta$ & $\delta$ & Direction \\
\midrule
1  & 300 & 5.00 & 1.00 & 11 & 20.0 & 0.00 & Upward   \\
2  & 150 & 5.00 & 1.00 & 12 & 40.0 & 0.00 & Upward   \\
3  & 500 & 5.00 & 1.50 & 13 & 0.0  & 1.50 & -     \\
4  & 100 & 5.00 & 0.75 & 14 & 0.0  & 3.00 & -     \\
5  & 100 & 5.00 & 0.60 & 15 & 20.0 & 1.50 & Upward   \\
6  & 450 & 8.00 & 2.00 & 16 & 40.0 & 1.50 & Upward   \\
7  & 450 & 6.00 & 2.00 & 17 & 20.0 & 3.00 & Upward   \\
8  & 100 & 8.00 & 1.00 & 18 & 40.0 & 3.00 & Upward   \\
9  & 150 & 10.0 & 2.00 & 19 & 30.0 & 2.25 & Sideways \\
10 & 100 & 10.0 & 1.50 & 20 & 30.0 & 2.25 & Downward \\
\botrule
\end{tabular*}  
\footnotetext[1]{From left to right: the geometry number, the maximum inlet velocity ($v_{max}$) in mms$^{-1}$, the inlet radius (R$_I$) in mm, and the constricted section radius (R$_C$) in mm.}
\footnotetext[2]{From left to right: the geometry number, the angle between the constricted section and the direction normal to the inlet surface ($\theta$) in degrees, the offset between the constricted section and the centre of the cylinder ($\delta$) in mm, and the direction that the constricted section is angled.}
\end{minipage}
\end{center}
\end{table}

\subsubsection{Boundary Conditions}
The relevant boundary conditions are those relating to the inlet, outlet, and inner wall. The inlet boundary conditions take the most work to define as the velocity varies both in space and time. For blood flowing through the AV, the flow is expected to be fully developed, that is, the velocity is zero at the inner walls due to friction and at its highest in the centre. This can be achieved by extending the length that the blood needs to travel from the inlet to the constricted section, allowing the flow to fully develop before reaching the aortic valve. However, extending the length means the geometry becomes larger, hence increasing the computational time. To compensate for this, the velocity profile at the inlet was defined with a parabolic shape and the upstream length was set to 20mm \cite{Madhavan2016}. This means the flow will start out more developed than with a uniform profile, and become fully developed before reaching the aortic valve. The velocity was also time-dependent and represented the diastole (where regurgitation occurs), with a rapid initial increase in magnitude before slowly decreasing \cite{Sallach2007}.

The parabolic velocity profile at the inlet in Cartesian coordinates was defined by 

\begin{equation}
    v = v_{max} - \frac{x^2+y^2}{R_I^2v_{max}^{-1}}
\end{equation} where $v_{max}$ is the maximum velocity at the centre of the inlet and R$_I$ is the inlet radius. The remaining boundary conditions were for the outlet and walls. The outlet was defined as an opening with zero pressure difference and the walls were defined as non-permeable with no-slip boundary conditions.

\subsection{Data Preparation}
To start, the raw CFD simulations and geometries, which each had 71 timeframes, were sampled onto a uniform Cartesian grid to be used as HR images. This was completed with linear interpolation, and took from 10 minutes to 3 hours per time frame depending on the size of the geometry and the CFD data. To separate the fluid and non-fluid regions for better data processing and result quantification, binary masks were generated using k-Nearest-Neighbours \cite{knearestneighbours}. The main difference between these synthetic HR images and MR images relates to the voxel size, VENC, and the amount of noise – HR images are noise-free whereas MR contain phase noise. The LR MR images were obtained from the HR images using the same method as in 4DFlowNet \cite{4DFlowNet} to simulate 4x downsampled MR images with appropriate noise and VENC. This gives the paired LR and HR synthetic images used in network training.

\subsection{Data Augmentation}
\begin{figure}
    \centering
    \includegraphics[width=\textwidth]{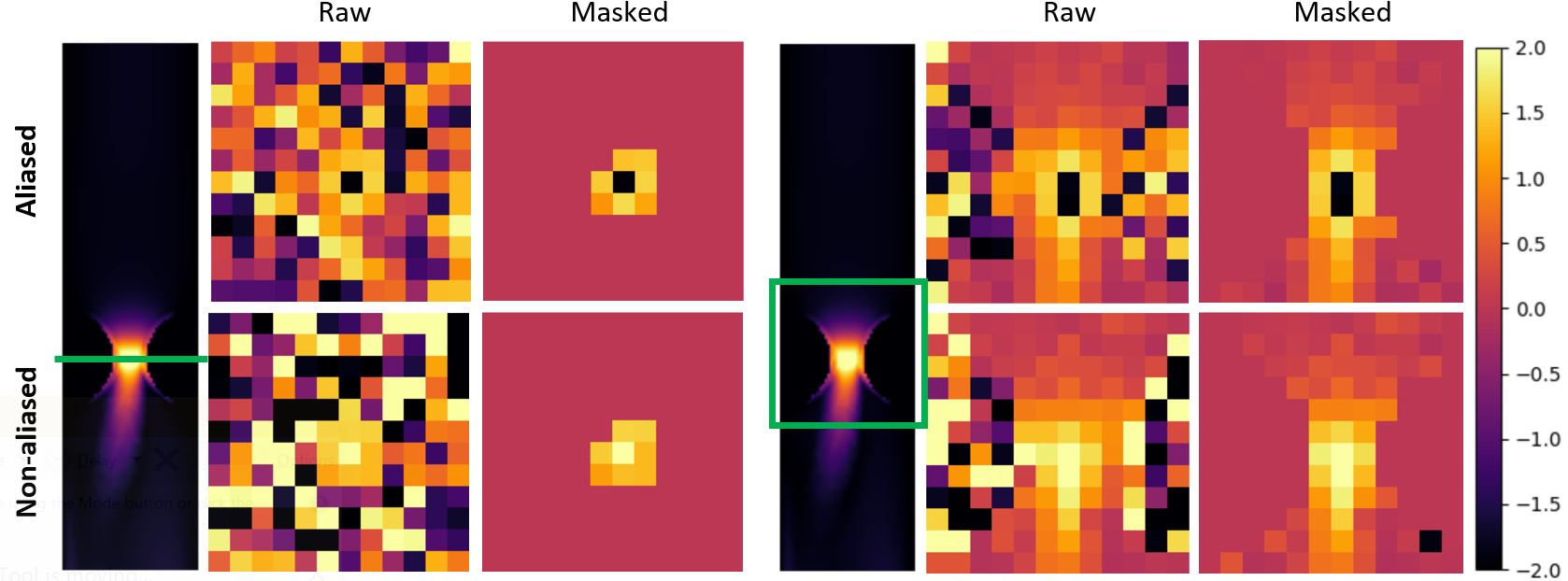}
    \caption{Examples of aliased patches against their non-aliased counterparts. 2D slices of the velocity in the $x$ direction were taken from the 3D patch for visualisation purposes, along the width (left) and length (right) of the geometry. Scale is in metres per second.}
    \label{fig:aliasing}
\end{figure}

To augment the data set, similar techniques were used as in 4DFlowNet \cite{4DFlowNet} for each time frame. VENC values were randomly chosen from a set of velocities between 0.3ms$^{-1}$ and 6.0ms$^{-1}$, spaced by 0.3ms$^{-1}$, for each velocity component. Aliasing was mostly avoided by choosing a VENC larger than the peak velocity. However, since velocity jets cannot be estimated beforehand for actual aortic regurgitation cases (which may cause phase aliasing), a VENC lower than the maximum velocity was chosen with a 10\% probability, randomly selecting between 0.3ms$^{-1}$ or 0.6ms$^{-1}$ lower. Some of these aliased patches are shown in Figure \ref{fig:aliasing}. Constant intensity values between 60 and 240 were randomly chosen for the magnitude image, and noise levels were added depending on the signal-to-noise ratio, which were randomly and uniformly chosen between 14 and 17 decibels.

Since there were a limited number of geometries, further augmentation came in patch generation. From each time frame, 10 patches of 12x12x12-voxel cubes from the LR image were selected randomly with a minimum fluid region of 20\%. These patches acted as random translations, so no extra translation steps were taken. On top of this, for each patch generated another randomly rotated version of the patch was also created. This resulted in 20 patches generated from each time frame and thus 1420 patches per geometry. 

\subsection{Training and Validation}
To investigate the effect of additional geometries regarding SR image quality, a subset of the data consisting of patches from only five geometries was compared against a the entire data set consisting of patches from all geometries. The validation set in both cases was the same, consisting only of patches from a single geometry.

To investigate the effect of aliasing, duplicates of the two training and validation sets were generated, but with a 10\% probability of having a VENC lower than the maximum velocity in any time frame. Networks trained using these aliased data sets were validated against the previously generated validation set without any aliasing, as well as a newly generated validation set with full aliasing, that is, with each time frame having a VENC lower than the maximum velocity.

\subsection{Network Architecture}
The simulated pairs of LR and HR synthetic images were used to train a similar deep residual network structure to the one in 4DFlowNet. This consisted of several residual blocks surrounding a central upsampling layer, with the preceding blocks in the LR space pre-processing and acting as denoisers for the input while the following blocks in the HR space refine the output. In 4DFlowNet, LR patches of 16-voxel cubes were used as input and SR patches of 32-voxel cubes were generated as output, with an upsample factor of 2.

Several changes were made to the above 4DFlowNet architecture to provide higher resolution images with improved accuracy. Firstly, the upsample factor was increased to 4 and the sizes of the input and output patches were changed to 12-voxel and 48-voxel cubes, respectively. The smaller patches account for smaller vessel sizes in the cardiovascular space around the aortic valve \cite{Ferdian-Cerebrov}. Secondly, the dense and cross stage partial blocks in DenseNet and CSPNet, respectively, were experimented with by using them in place of the 12 residuals blocks in the original 4DFlowNet architecture. The growth rate \cite{DenseNet}, defined as the number of feature maps in each convolutional layer, of the dense and CSP blocks was set to 16, a quarter of the number of channels in each convolutional layer from the original residual blocks. The adapted 4DFlowNet architecture with residuals blocks (4DFlowNet-Res), with dense blocks (4DFlowNet-Dense), and with cross stage partial blocks (4DFlowNet-CSP) had 3.34, 2.55, and 2.08 million parameters, respectively. These modified networks were implemented with TensorFlow 2.0 \cite{Tensorflow20} and trained using an Adam optimiser \cite{kingma2017adam}, with an initial learning rate of $10^{-4}$ and decay rate of $\sqrt{2}$ after every 14 epochs. Batch sizes of 16 were used, with training completed in 200 epochs.

\subsection{Loss Function}
The network was optimised by minimising the mean squared error (MSE) between the paired HR images and the SR images generated from the corresponding input LR ones. The voxel-wise loss was calculated as the mean sum of squared differences between each velocity component:

\begin{equation}
    L_{MSE} = \frac{1}{N} \sum^N_{i = 1}{(v'_{x_i} - v_{x_i})^2 + (v'_{y_i} - v_{y_i})^2 + (v'_{z_i} - v_{z_i})^2}
\end{equation} where $N$ is the total number of voxels in the geometry, $v'_j$ is the predicted SR velocity, and $v_j$ is the actual HR velocity, for $j \in \{x, y, z\}$.

The MSE of fluid and non-fluid regions were calculated as separate terms due to the imbalance and irregularity of these regions within a specific patch. This gives the total loss to be:

\begin{equation}
    L_{total} = L_{MSE_{F}} + L_{MSE_{N}}
\end{equation} where $L_{MSE_{F}}$ and $L_{MSE_{N}}$ are the voxel-wise loss for the fluid and non-fluid regions, respectively. 

The original loss function in 4DFlowNet contained a weighted velocity gradient term to smoothen the gradient between neighbouring vectors \cite{4DFlowNet}. This was omitted from the above loss function as improvements were observed in near-wall velocity estimates with its removal \cite{Ferdian-Cerebrov}.

\subsection{Evaluation Metric}
The relative speed error (RE), the relative difference between the SR velocity magnitude (speed) compared to the actual HR speed on the validation set, was used to measure network performance and save model checkpoints. This was only calculated in fluid regions to avoid zero division error, as well as adding a small number ($\epsilon = 10^{-4}$) to the denominator. Furthermore, since many speed values in the HR images were quite small, this could risk significantly over-penalising the model. Thus, an arctangent approach \cite{MAAPE} was adopted, giving the following equation for relative speed error:

\begin{equation}
    RE = \frac{1}{N} \sum^N_{i = 1}arctan(\frac{\sqrt{(v'_{x_i} - v_{x_i})^2 + (v'_{y_i} - v_{y_i})^2 + (v'_{z_i} - v_{z_i})^2}}{\sqrt{v_{x_i}^2 + v_{y_i}^2 + v_{z_i}^2} + \epsilon})
\end{equation} where $N$ is the total number of voxels in the fluid domain, $v'_j$ and $v_j$ are the predicted SR and actual HR velocities, respectively, for all $j \in \{x, y, z\}$, and `arctan' is the arctangent function, defined for all real values from negative infinity to infinity with $lim_{x\rightarrow{\infty}}tan^{-1}x=\frac{\pi}{2}$ for $arctan$ $x$.

In addition to the RE, network performance was also evaluated using the root mean squared error (RMSE) and the structural similarity (SSIM) metric \cite{SSIM} in all three Cartesian velocity components. These were compared against the baseline 4DFlowNet model that had been trained with an upsample factor of 2.

\newpage
\section{Results}\label{sec4}
\begin{figure}
    \centering
    \includegraphics[width=\textwidth]{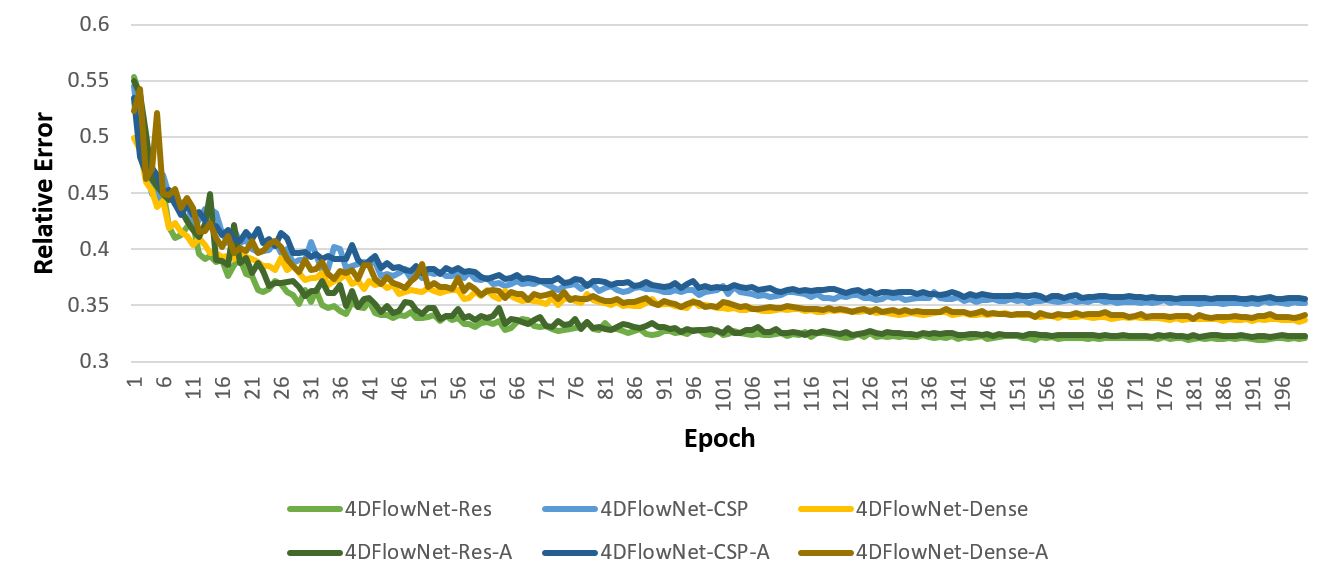}
    \caption{Relative error across all 200 epochs during training for each network. Networks trained with five geometries were not included as the general trend seen was the same.}
    \label{fig:training}
\end{figure}

Training was performed using a Tesla V100 GPU with 32GB memory with networks being trained for 200 epochs. Improvements in relative speed error (RE) plateaued around the 100 epoch mark for 4DFlowNet-Res while still improving for 4DFlowNet-CSP and 4DFlowNet-Dense up till the very last epoch. This can be seen in Figure \ref{fig:training}. The time taken was dependent on the type of network; 4DFlowNet-Res, 4DFlowNet-CSP, and 4DFlowNet-Dense took approximately 163, 168, and 255 hours, respectively. Note that these times were for the networks trained using all geometries. For the networks trained using only five geometries, denoted as 4DFlowNet-Res5, 4DFlowNet-CSP5, and 4DFlowNet-Dense5, the times taken were approximately 38, 40, and 63 hours, respectively. There was no significant difference in training time between networks trained with and without a portion of aliased data, denoted by `-A'.

Networks were tested on one complete geometry consisting of 71 timeframes with no (phase/velocity) aliasing, and the same complete geometry with full aliasing. These predictions were required to be patch-based since patches were used as the input and output for each network. The complete geometry was reconstructed by stitching together multiple SR velocity field patches, which was done with a stride of ($n-4$) in each Cartesian direction where n is the arbitrary patch size. To avoid patch artifacts at the boundary, four voxels were stripped from each patch side. 

\subsection{Synthetic MR Images}

\begin{figure}
    \centering
    \includegraphics[width=\textwidth]{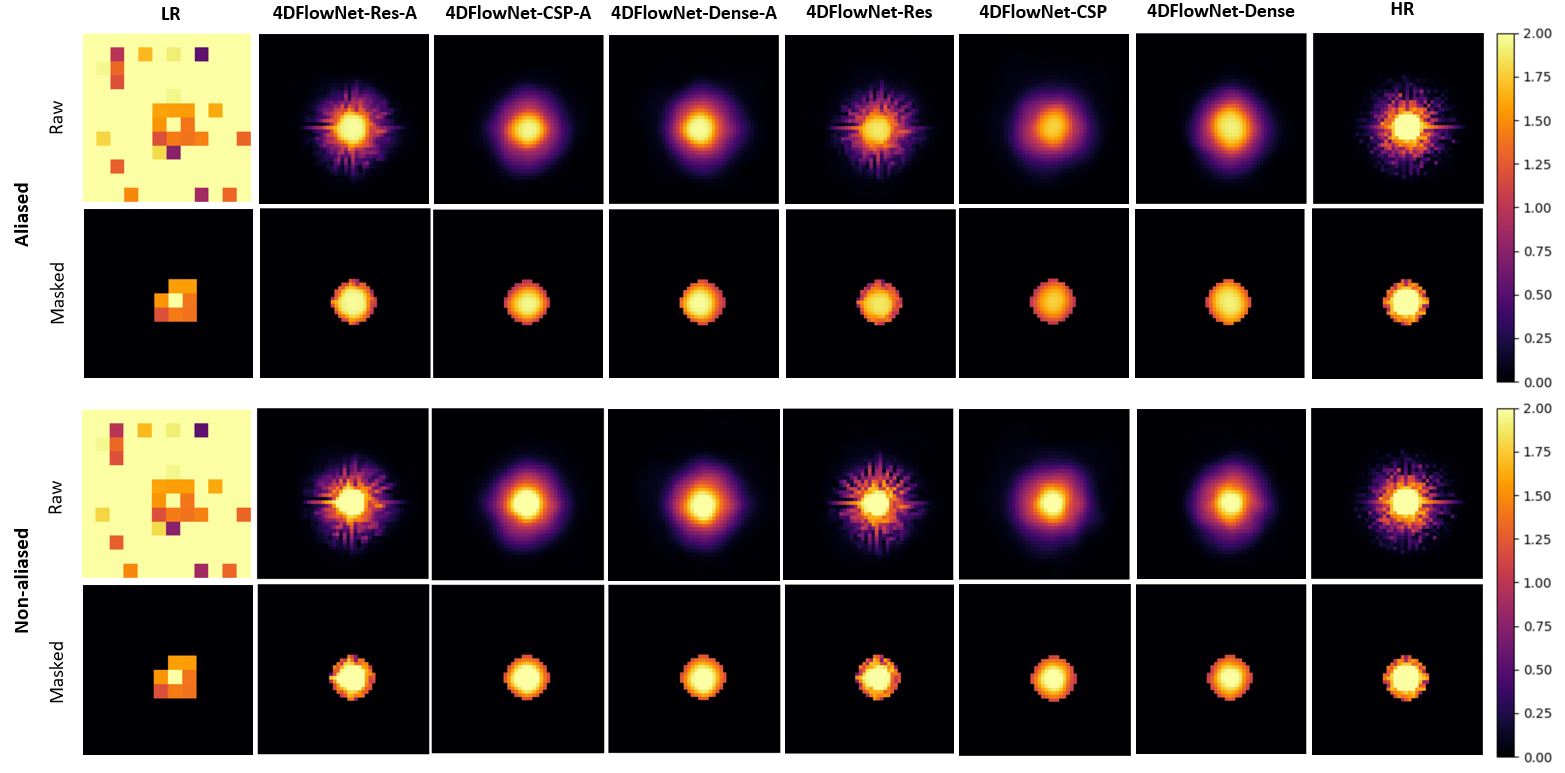}
    \caption{Predictions on an LR patch in the constricted region from the synthetic 4D flow MRI phase image for different networks, with (top) and without (bottom) aliasing error. A 2D slice, along the width of the geometry, of the velocity magnitude is shown from the 3D patch for visualisation purposes. Scale is in metres per second.}
    \label{fig:pred_visual1}
\end{figure}

\begin{figure}
    \centering
    \includegraphics[width=\textwidth]{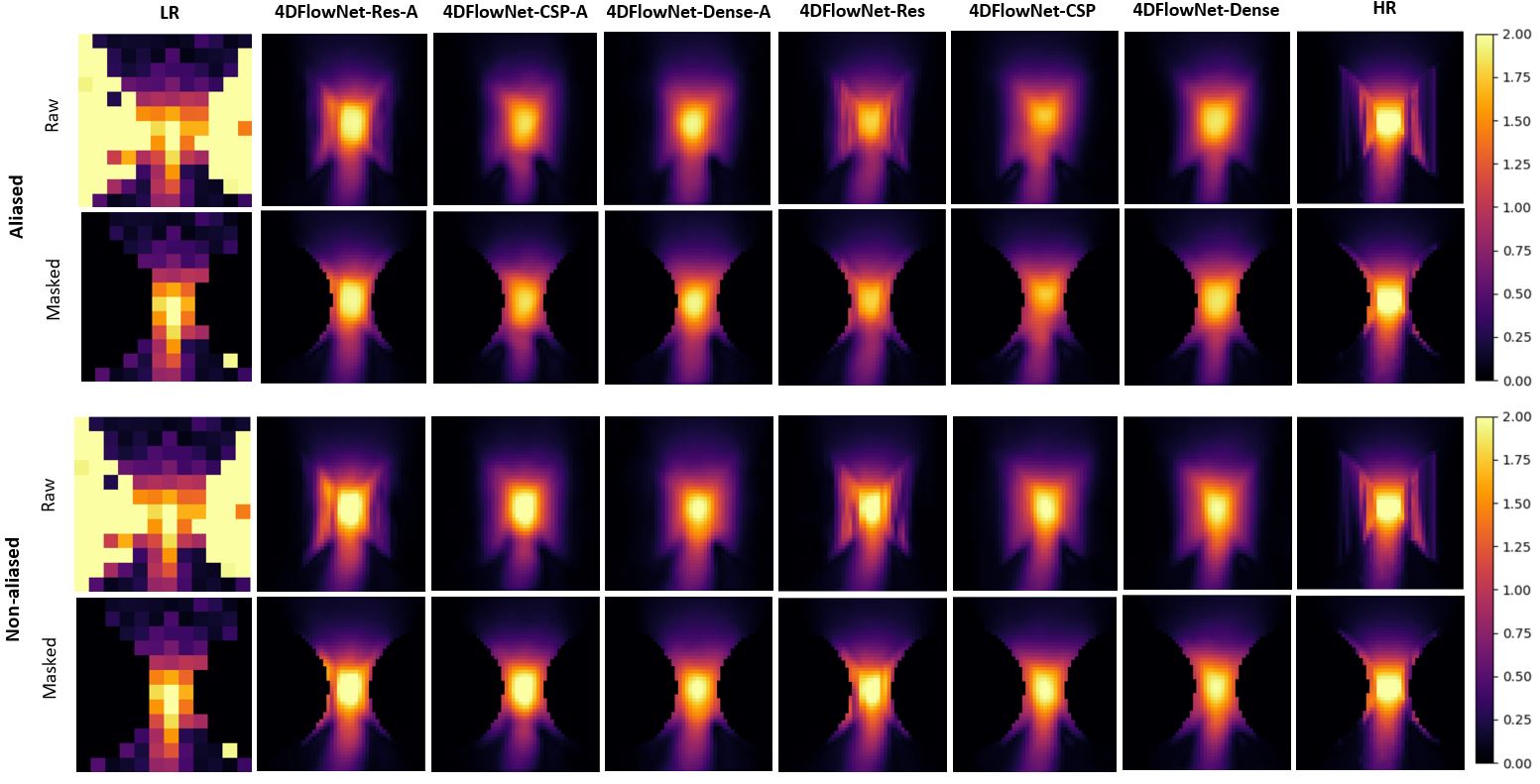}
    \caption{Predictions on an LR patch, focused on the constricted section, from the synthetic 4D flow MRI phase image for different networks, with (top) and without (bottom) aliasing error. A 2D slice, along the length of the geometry, of the velocity magnitude is shown from the 3D patch for visualisation purposes. Scale is in metres per second.}
    \label{fig:pred_visual2}
\end{figure}

SR images were analysed visually and quantitatively to better understand how each model was performing. Figures \ref{fig:pred_visual1} and \ref{fig:pred_visual2} are visual examples of the prediction for the different networks in the constricted section at the peak flow. These display the effectiveness of each network in reducing noise, with the predictions looking quite similar to the ground truth. Furthermore, the networks trained with a proportion of aliased data seem to be performing better than networks without, especially for data with aliasing error. 

The values for each evaluation metric were collected and compared in Tables \ref{NA_tab} and \ref{A_tab} to quantify the performance of every model. Again, these values were taken from the time frame with peak flow, with peak velocities of 2.186, 0.355, and 0.349 ms$^{-1}$ for velocity components $v_x$, $v_y$, and $v_z$, respectively. 

\begin{sidewaystable}
\sidewaystablefn%
\begin{center}
\begin{minipage}{\textheight}
\caption{Summary of prediction errors and evaluation metrics for different networks, when predicting on non-aliased data.}\label{NA_tab}
\begin{tabular*}{\textheight}{@{\extracolsep{\fill}}lccccc@{\extracolsep{\fill}}}
\toprule%%
Network & RMSE$_x$\footnotemark[1] & RMSE$_y$\footnotemark[1] & RMSE$_z$\footnotemark[1] & SSIM\footnotemark[2] & RE\footnotemark[3] \\
\midrule
4DFlowNet          & 0.0574 ± 0.0574 & 0.0198 ± 0.0198 & 0.0177 ± 0.0177 & (0.832, 0.708, 0.705) & 0.543 \\
4DFlowNet-Res      & 0.0361 ± 0.0360 & 0.0138 ± 0.0138 & 0.0135 ± 0.0135 & (0.923, \textbf{0.843}, 0.820) & 0.330 \\
4DFlowNet-CSP      & 0.0354 ± 0.0346 & 0.0142 ± 0.0142 & 0.0154 ± 0.0153 & (0.915, 0.728, 0.766) & 0.316 \\
4DFlowNet-Dense    & 0.0372 ± 0.0359 & 0.0138 ± 0.0138 & 0.0139 ± 0.0138  & (0.908, 0.794, 0.789) & \textbf{0.296} \\
4DFlowNet-Res-A    & 0.0369 ± 0.0363 & 0.0136 ± 0.0136 & 0.0138 ± 0.0136 & (0.924, 0.829, 0.824) & 0.340 \\
4DFlowNet-CSP-A    & 0.0366 ± 0.0366 & \textbf{0.0135 ± 0.0135} & \textbf{0.0126 ± 0.0124} & (\textbf{0.939}, 0.812, 0.803) & 0.330 \\
4DFlowNet-Dense-A  & \textbf{0.0350 ± 0.0343}  & 0.0149 ± 0.0148 & 0.0163 ± 0.0163 & (0.912, 0.747, 0.705) & 0.301 \\
4DFlowNet-Res5     & \textcolor{red}{0.0375 ± 0.0375} & 0.0160 ± 0.0158  & 0.0163 ± 0.0162 & (0.934, 0.796, \textcolor{red}{\textbf{0.825}}) & \textcolor{red}{0.344} \\
4DFlowNet-CSP5     & 0.0399 ± 0.0397 & \textcolor{red}{0.0152 ± 0.0152} & \textcolor{red}{0.0145 ± 0.0142}  & (\textcolor{red}{0.935}, 0.812, 0.804) & 0.338 \\
4DFlowNet-Dense5   & 0.0407 ± 0.0406 & 0.0166 ± 0.0166  & 0.0150 ± 0.0147 & (0.929, 0.764, 0.799) & 0.352 \\
4DFlowNet-Res5-A   & 0.0415 ± 0.0415 & 0.0196 ± 0.0196 & 0.0189 ± 0.0189 & (0.924, \textcolor{red}{0.815}, 0.817) & 0.354 \\
4DFlowNet-CSP5-A   & 0.0438 ± 0.0438 & 0.0219 ± 0.0219 & 0.0211 ± 0.0211 & (0.929, 0.803, 0.811) & 0.362 \\
4DFlowNet-Dense5-A & 0.0402 ± 0.0402 & 0.0194 ± 0.0193 & 0.0169 ± 0.0169 & (0.928, 0.807, 0.784) & 0.364 \\
\botrule
\end{tabular*}
\footnotetext{Note: The best values for each metric are in \textbf{bold}. The best values for networks trained with only five geometries have also been coloured \textcolor{red}{red}.}
\footnotetext[1]{RMSE$_i$ ± s.d. for velocity component $v_i$, in metres per second.}
\footnotetext[2]{Includes all three Cartesian velocity components in the form ($v_x$, $v_y$, $v_z$).}
\footnotetext[3]{Or equivalently, MAAPE.}
\end{minipage}
\end{center}
\end{sidewaystable}

\begin{sidewaystable}
\sidewaystablefn%
\begin{center}
\begin{minipage}{\textheight}
\caption{Summary of prediction errors and evaluation metrics for different networks, when predicting on aliased data.}\label{A_tab}
\begin{tabular*}{\textheight}{@{\extracolsep{\fill}}lccccc@{\extracolsep{\fill}}}
\toprule%%
Network & RMSE$_x$\footnotemark[1] & RMSE$_y$\footnotemark[1] & RMSE$_z$\footnotemark[1] & SSIM\footnotemark[2] & RE\footnotemark[3] \\
\midrule
4DFlowNet          & 0.0642 ± 0.0641 & 0.0292 ± 0.0292 & 0.0272 ± 0.0272 & (0.855, 0.724, 0.814) & 0.512 \\
4DFlowNet-Res      & 0.0361 ± 0.0361 & \textbf{0.0142 ± 0.0142} & \textbf{0.0138 ± 0.0138} & (0.927, 0.858, 0.836) & 0.334 \\
4DFlowNet-CSP      & 0.0364 ± 0.0360 & 0.0161 ± 0.0161 & 0.0164 ± 0.0163 & (0.917, 0.795, 0.778) & 0.309 \\
4DFlowNet-Dense    & 0.0356 ± 0.0344 & 0.0153 ± 0.0153 & 0.0159 ± 0.0159 & (0.917, 0.812, 0.798) & 0.309 \\
4DFlowNet-Res-A    & 0.0355 ± 0.0350 & 0.0148 ± 0.0148 & 0.0140 ± 0.0138 & (0.925, 0.845, 0.848) & 0.342 \\
4DFlowNet-CSP-A    & 0.0352 ± 0.0352 & 0.0158 ± 0.0158 & 0.0147 ± 0.0146 & (\textbf{0.942}, 0.845, \textbf{0.849}) & 0.313 \\
4DFlowNet-Dense-A  & \textbf{0.0340 ± 0.0343} & 0.0164 ± 0.0148 & 0.0164 ± 0.0163 & (0.922, 0.785, 0.757) & \textbf{0.284}  \\
4DFlowNet-Res5     & 0.0468 ± 0.0468 & 0.0176 ± 0.0176 & 0.0170 ± 0.0169 & (0.865, 0.899, 0.582) & \textcolor{red}{0.337} \\
4DFlowNet-CSP5     & 0.0529 ± 0.0529 & \textcolor{red}{0.01743 ± 0.01743} & 0.01636 ± 0.01619 & (0.912, 0.921, 0.509) & 0.341 \\
4DFlowNet-Dense5   & 0.0508 ± 0.0508 & 0.0178 ± 0.0177 & 0.0171 ± 0.0171 & (0.902, 0.910, 0.587) & 0.353 \\
4DFlowNet-Res5-A   & \textcolor{red}{0.0437 ± 0.0437} & 0.0175 ± 0.0175 & \textcolor{red}{0.0159 ± 0.0159} & (0.901, 0.902, 0.522) & 0.344 \\
4DFlowNet-CSP5-A   & 0.0466 ± 0.0465 & 0.0180 ± 0.0180 & 0.0164 ± 0.0164 & (\textcolor{red}{0.914}, \textcolor{red}{\textbf{0.925}}, \textcolor{red}{0.610}) & 0.350  \\
4DFlowNet-Dense5-A & 0.0474 ± 0.0473 & 0.0194 ± 0.0193 & 0.0162 ± 0.0162 & (0.895, 0.902, 0.506) & 0.338 \\
\botrule
\end{tabular*}
\footnotetext{Note: The best values for each metric are in \textbf{bold}. The best values for networks trained with only five geometries have also been coloured \textcolor{red}{red}.}
\footnotetext[1]{RMSE$_i$ ± s.d. for velocity component $v_i$, in metres per second.}
\footnotetext[2]{Includes all three Cartesian velocity components in the form ($v_x$, $v_y$, $v_z$).}
\footnotetext[3]{Or equivalently, MAAPE.}
\end{minipage}
\end{center}
\end{sidewaystable}

\begin{figure}
    \centering
    \includegraphics[width=0.8\textwidth]{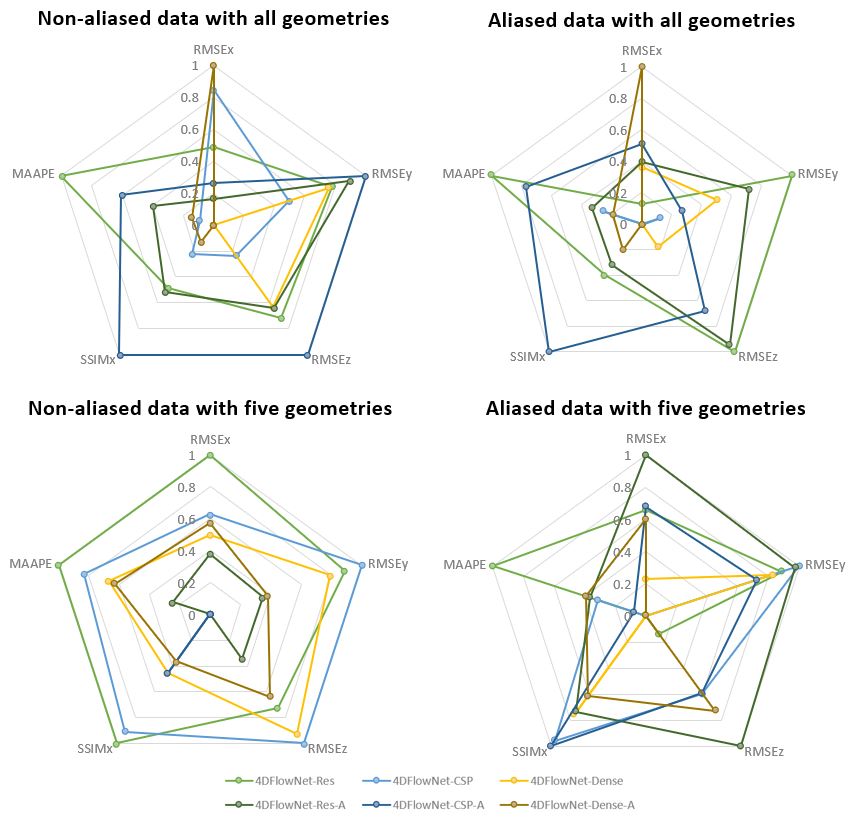}
    \caption{Comparison of networks for different metrics and results, varied by the data set size and whether aliasing was used. Values were normalised to be within 0 and 1, representing the worst and best values, respectively, for a specific metric relative to other networks.}
    \label{fig:radar}
\end{figure}

The metrics were plotted in Figure \ref{fig:radar}. Briefly, 4DFlowNet-Dense-A and 4DFlowNet-CSP-A seem to perform the best with the lowest RMSE error and largest SSIM in the principle flow direction ($v_x$), respectively, on both aliased and non-aliased data. However, there does appear to be considerable variation in these results between different networks, depending heavily on the size of the data set and whether aliasing is present. 

Regarding the RMSE in the principle flow direction (RMSE$_x$), all networks perform better, on average, than the base 4DFlowNet with less variation in RMSE$_x$. For the smaller data set with five geometries, the residual-based (Res) versions seem to have the smallest RMSE$_x$. There is also noticeable improvement in error between networks trained on the smaller and larger datasets when predicting on non-aliased data. However, when predicting on aliased data, the improvement is significantly more apparent. The increase in dataset size improves the RMSE in the other two flow directions for all networks too, with the CSP versions somewhat better than other versions. 

Regarding the SSIM in the principle flow direction (SSIM$_x$), all networks also perform better, on average, than the base 4DFlowNet. However, the SSIM$_x$ seems to worsen for a larger dataset, in general, when predicting on non-aliased data. On the other hand, when predicting on aliased data, there does appear to be slight improvement in SSIM$_x$ across all networks. For the SSIM in the other two flow directions, the values are considerably more varied and worse than in the principle direction. Finally, all networks have a substantially lower RE than the base, with the Dense versions having the lowest RE.

\begin{figure}
    \centering
    \includegraphics[width=\textwidth]{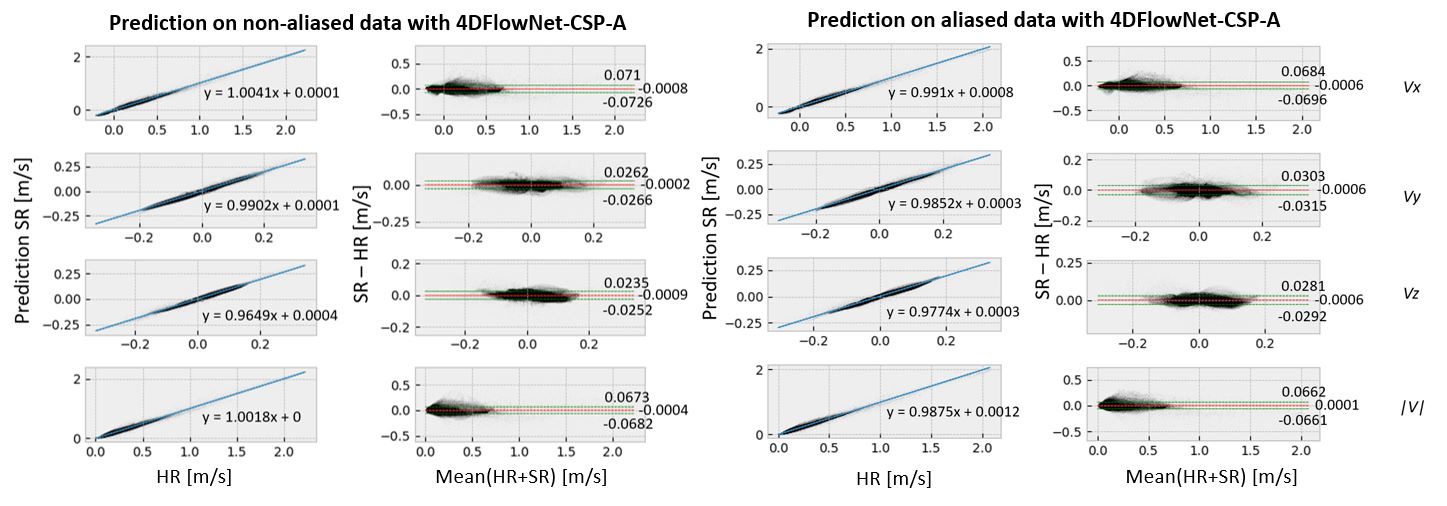}
    \caption{Regression and Bland-Altman plots for prediction on non-aliased (left) and aliased (right) data with 4DFlowNet-CSP-A. These plots are for each of the velocity components and magnitude ($v_x$, $v_y$, $v_z$, and $\|v\|$, respectively, from top to bottom) between SR and synthetic HR images.}
    \label{fig:pred_CSP-A}
\end{figure}

\begin{figure}
    \centering
    \includegraphics[width=\textwidth]{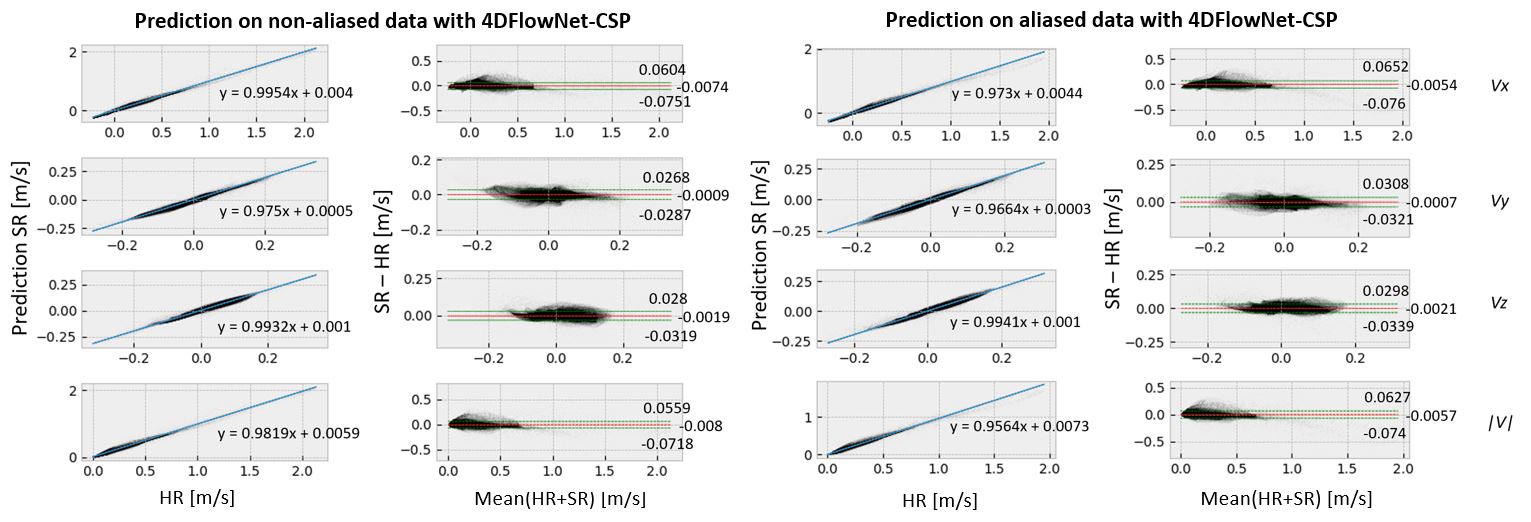}
    \caption{Regression and Bland-Altman plots for prediction on non-aliased (left) and aliased (right) data with 4DFlowNet-CSP. These plots are for each of the velocity components and magnitude ($v_x$, $v_y$, $v_z$, and $\|v\|$, respectively, from top to bottom) between SR and synthetic HR images.}
    \label{fig:pred_CSP}
\end{figure}

The regression plots in Figures \ref{fig:pred_CSP-A} and \ref{fig:pred_CSP} show that there is exceptional correlation between the SR and synthetic HR images. The regression slopes are very close to one, in the principle flow direction and velocity magnitude plots, for the two CSP networks. Moreover, offset values are also essentially zero in all examples shown. Training with all non-aliased images appears to have an effect when predicting the higher velocity values in aliased images, with these values being slightly underestimated, as seen on the right in Figure \ref{fig:pred_CSP} and confirmed by the Bland-Altman plots too. These plots also indicate minimal bias as the deviations appear constant, uniform, and are all less than 0.08ms$^{-1}$.

\begin{figure}
    \centering
    \includegraphics[width=\textwidth]{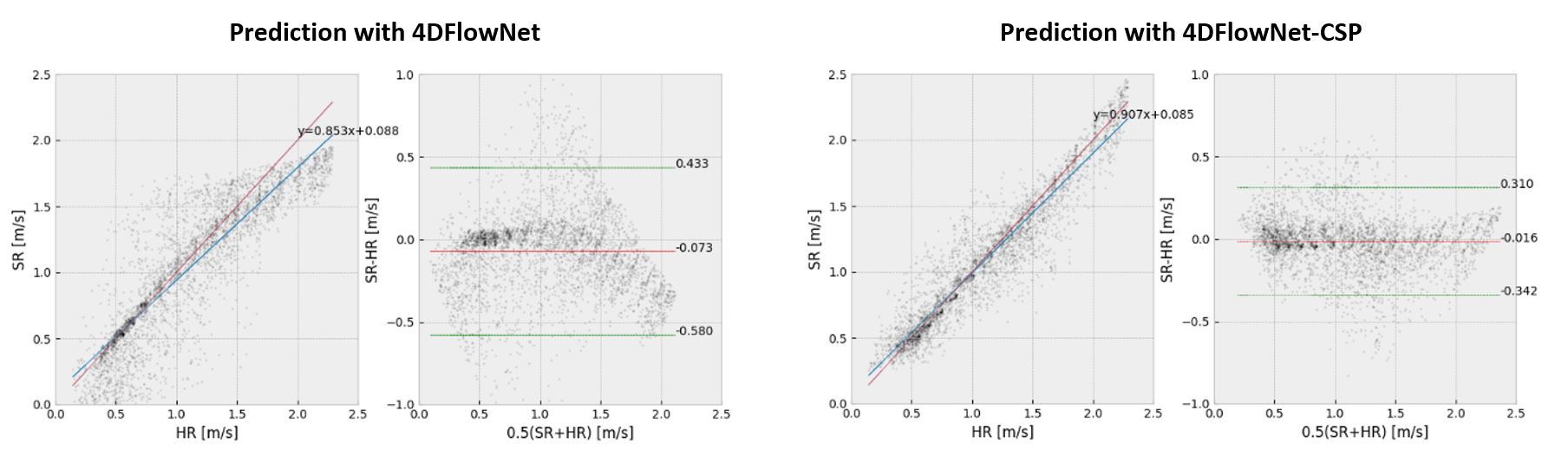}
    \caption{Regression and Bland-Altman plots for $\|v\|$ prediction within the constricted section, comparing the baseline 4DFlowNet (left) against the adapted 4DFlowNet-CSP (right).}
    \label{fig:RBA-stenosis2}
\end{figure}

For velocities within the constricted section, Figure \ref{fig:RBA-stenosis2} shows the correlation between SR and synthetic HR images. Although a slight underestimation bias seem to prevail, 4DFlowNet-CSP shows noticeable improvement from the baseline 4DFlowNet improving on the otherwise observed deviations at higher velocities. Note that the general trends and comments regarding these regression and Bland-Altman plots were present in all other evaluated networks too.

\begin{figure}
    \centering
    \includegraphics[width=\textwidth]{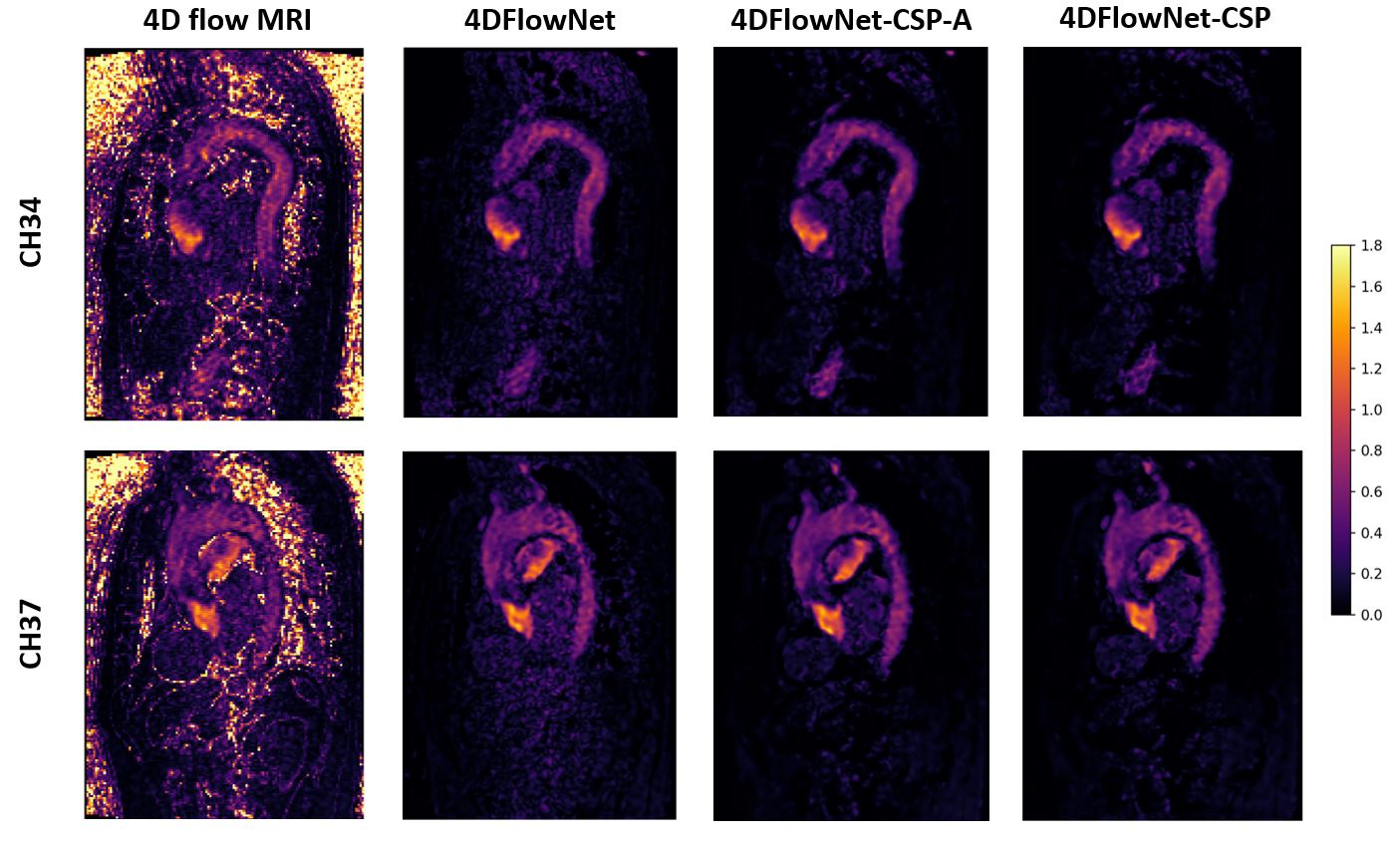}
    \caption{Predicted SR images on actual 4D flow MRI data. Two sets of 4D flow MRI data were used (top and bottom), with the original LR images on the left and the SR images in the other three columns, titled by the network used for prediction. These are 2D slices of the 3D image, showing the velocity magnitude taken at peak flow. Scale is in metres per second.}
    \label{fig:mri_pred}
\end{figure}

\subsection{In-vivo 4D Flow MRI Data}

Ethical approval for this study was granted by the Health and Disability Ethics Committee of New Zealand (17/CEN/226), and written informed consent was obtained from each participant. Two sets of in-vivo 4D flow MRI data (CH34 and CH37) were acquired and used to show how the network would perform, seen in Figure \ref{fig:mri_pred}. These were only LR images, with no HR images available to compare the predicted SR images against. However, the SR images produced were compared against the baseline 4DFlowNet to help better understand the predictions. The SR images seem to have effectively removed noise from the LR image, with the new 4DFlowNet-CSP networks appearing to remove slightly more noise throughout the LR image than the baseline 4DFlowNet. Other networks performed similarly, with predicted SR images almost identical to the ones shown. 

\begin{figure}
    \centering
    \includegraphics[width=\textwidth]{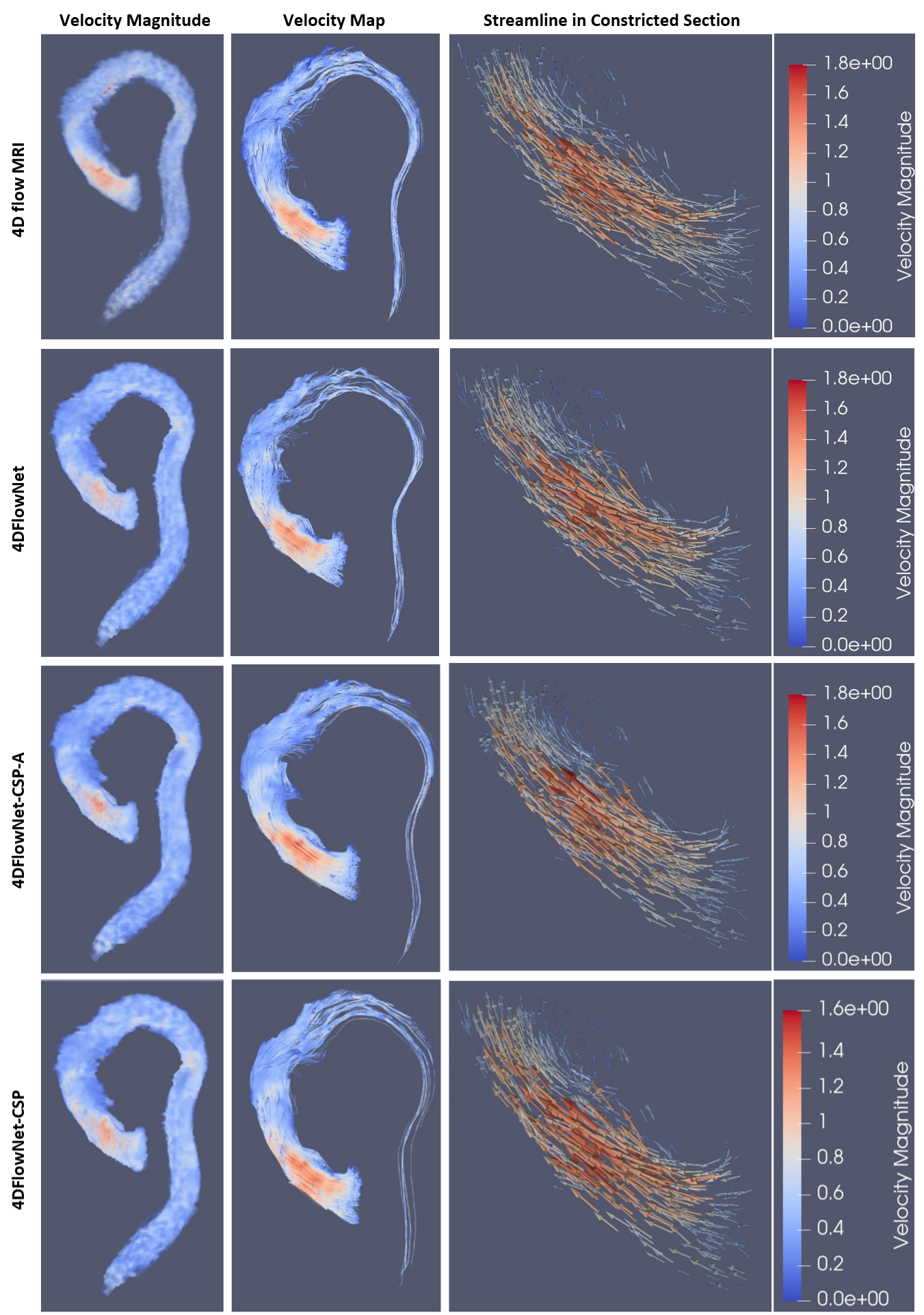}
    \caption{Predicted SR images visualised in ParaView. The columns show the velocity magnitude, its vector field, and the streamline reconstruction within the largest magnitde section. Scale is in metres per second.}
    \label{fig:mri_pred_paraview}
\end{figure}

The top set of data (CH34) was also segmented and visualised with ParaView \cite{paraview}, shown in Figure \ref{fig:mri_pred_paraview}. Image stitching appears to have been performed correctly, with velocities within the fastest section preserved well.

\newpage

\section{Discussion}\label{sec5}

A noteworthy consideration when analysing the results is that the focus should be more towards metrics and values obtained in the principle flow direction $x$. Since the peak velocities in the other flow directions, $y$ and $z$, were almost 10 times smaller than that in the principle direction, networks would have difficulty differentiating between velocity and noise. This is evident in the $y$ and $z$ RMSE values, which were over half of their corresponding peak velocities. The $y$ and $z$ SSIM values were also much worse, potentially exhibiting strange behaviour in these low velocity fields too \cite{SSIM_limitations}. 

\subsection{Synthetic 4D Flow MRI Data}

The main limitation of previous work has been related to insufficient data and flow characteristics \cite{4DFlowNet, Cerebrovascular-4DFlowMRI, Ferdian-Cerebrov}. To understand the effect of additional flow characteristics in the dataset, a wider range was used in this study. There was definitely noticeable improvement in the RMSE across all networks tested, particularly when predicting on aliased data. This indicates that with more geometries and hence flow characteristics, the network seems to generalise better and is more robust against data it has not seen.

In terms of the synthetic LR and HR image pairs, the downsampling process and patch-based approach seemed to work effectively. The noisy LR patches enabled the network to learn noise removal while also enabling greater generalisation to unknown flow characteristics or geometries \cite{4DFlowNet}. However, a small portion of these incorrect non-fluid velocity areas around the fluid domain appear to have been included in network training, seen in Figures \ref{fig:pred_visual1} and \ref{fig:pred_visual2}. These were generated during the linear interpolation process when obtaining the HR images from the CFD data. This suggests that the binary mask for separating between the fluid and non-fluid regions, created using k-Nearest-Neighbours, needs improvement. An obvious consequence is that the network may learn incorrect flow characteristics and predict less accurately. 

Incorporating aliased patches into the training data seemed to work effectively as well, improving the RMSE across networks trained with all geometries. This was done by choosing VENC values lower than the maximum velocity, within a particular time frame, with a 10\% probability. Note that this probability was chosen arbitrarily. For the VENC, if it is set too high, visualization of the jet may not be obtained and be inaccurate, as well as having poorer SNR. On the other hand, if it is set too low, flow characteristics may be lost and a mosaic pattern will be shown \cite{MRI_venc}. This means that for the time frames with aliasing, the velocities lower than the VENC would have been captured significantly more clearly, with only a small portion of high velocity characteristics being lost. For these time frames, the purpose would be to help the network better learn the flow characteristics in lower velocity fields, improving robustness and generalisability. However, this hypothesis has not been proven yet and more experiments on aliasing will be required to fully understand its effect on network performance. This would involve varying its probability of occurring as well as choosing VENC values considerably lower than the maximum velocity.

\subsection{Network Architecture}

A major drawback of residual blocks is that they suffer from limited learning ability \cite{DenseNet}. This was seen in the results, in which the residual networks did not show much improvement even after adding many more geometries or introducing aliased images into the training data. Furthermore, the RE for residual networks seemed to plateau at around the 100\textsuperscript{th} epoch, whereas the RE for the other networks seemed to continue improving, although at a decreasing rate, even up till the last epoch. These observations can be seen in Figure \ref{fig:training}. Despite this, the residual networks still performed well, with RMSE, SSIM, and RE values similar to the other networks, as well as having the fastest training and prediction times. This suggests that, when data is insufficient or very limited, residual networks may work best.

With sufficient or abundant data, cross stage partial or dense networks may be preferred. The learning ability in these types of networks have a significantly higher ceiling than residual networks \cite{DenseNet, CSPNet}, with the only difference between these two network structures being the training and prediction times. Cross stage partial networks had training and prediction times almost as fast as residual networks, whereas dense networks were almost 1.6 times slower. Although training times may not be a crucial problem, clinicians and patients alike may require and desire fast prediction times. This leads to cross stage partial networks being preferred. Otherwise, if training and prediction times are not significant constraints, then dense networks may be the optimal choice. Furthermore, the growth rate for the dense and cross stage partial networks was set quite low, at a quarter of the number of feature maps in each convolutional layer within the residual blocks. This limits the learning ability of these networks, as there are significantly less parameters, so testing larger values of this hyperparameter will be beneficial and likely improve network performance. Similarly, only a quarter of feature maps were taken from the base input layer within each partial dense block within the cross stage partial networks, so larger values of this hyperparameter will likely improve performance as well. 

Despite the network architecture seeming to work quite effectively, there are still improvements that could be made on top of modifying the residual blocks. Presently, the network is not taking full advantage of the temporal aspect in 4D flow MRI. Modifying it to incorporate characteristics of recurrent neural networks \cite{RecurrentNN} may help the network understand this temporal aspect better. This could be done by using predictions of the same patches from one or two time frames prior. Additionally, including physical properties of fluids may also help the network in learning flow characteristics, as seen in \cite{Physics-informed-4DFlowMRI, Ferdian-Cerebrov}. However, due to the bulky nature of the velocity data, which were 3D volumes for each velocity component, these ideas were not considered further as it would have been too costly to process given the available resources. 

\subsection{Clinical Application}

The clinical motivation for the current study was the assessment of  regurgitant or highly stenotic valvular flow; instances where quantification of regional velocities and changes in pressure act as effective biomarkers to describe the severity of disease. In these instances, and in clinical practice, assessment of \textit{peak} velocities through the vena contracta are used to symbolize disease severity. In fact, derivation of regional pressure drops are routinely derived from such peak velocity measures using the so called simplified Bernoulli equation \cite{stamm1983quantification} (coupling peak velocities to effective pressure changes). In this light, our results bear possible clinical implications in that SR velocities effectively recovers HR reference measures. This holds true throughout the evaluation domain, and in the clinically important constricted section, underestimation biases associated with the original 4DFlowNet formulation were effectively suppressed as seen in Figure \ref{fig:RBA-stenosis2}. The effect of the remaining minor deviation from a true 1:1 correlation between HR and SR data remains to be explored in a larger clinical setting, and in more complex flow scenarios - such as in the instance of regurgitant flow - higher-order methods might be required to derive pressure drops from measured velocity data \cite{marlevi2019estimation,marlevi2021noninvasive}. Nevertheless, just as we have shown the potential of recovering functional hemodynamic behavior through the spatially challenging cerebrovascular space using SR 4D Flow MRI \cite{Ferdian-Cerebrov}, the results of the current study bear similar potential in recovering clinically relevant hemodynamic metrics in aortic regurgitation. 

\subsection{Limitations}

Although the number of geometries and flow patterns have significantly increased from previous studies, a wider range of characteristics can still be included. Additional data will diversify the data set further and improve the model's robustness and generalisability even more. On top of this, more testing on real 4D flow MRI data will be required to validate model performance. Currently, this validation is only done by visually analysing model predictions to see if they look reasonable and sensible. The preferred approach would to be validate quantitatively with a pair of corresponding LR and HR 4D flow MRI images, calculating metrics such as RMSE and SSIM to properly understand model performance.

Lastly, several practical limitations can also be noted. Due to the limited GPU resources, training took approximately 60 days for all networks. With either more time or increased GPU resources, more networks could be trained by using different values for hyperparameters such as the growth rate and the probability that aliasing occurs. Moreover, networks could also be trained for more epochs, or until the error plateaus, to better gauge the learning ability of different networks. Finally, memory constraints were a factor too, as an upsample factor of 4 led to 64 times more usage in disk space. This would not be feasible for much larger upsample factors, so a different data representation may be required in future work. 

\section{Conclusion}

In this study, it was shown how deep learning and computational fluid dynamics can be combined to effectively quantify hemodynamics for aortic regurgitation. 4DFlowNet was successfully adapted and modified to produce 4D flow MRI SR images with an upsample factor of 4. The results show that by adding more geometries and hence flow characteristics into the data set, the accuracy of 4DFlowNet predictions are improved. Moreover, the comparison of different network architecture suggested that the original residual network structure limits learning ability and can be further refined.
\newpage
\bibliography{sn-bibliography}

\end{document}